\documentclass[conference,10pt]{IEEEtran}
\IEEEoverridecommandlockouts
\usepackage{amsmath,amssymb,amsfonts}
\usepackage{algorithmic}
\usepackage{graphicx}
\usepackage{textcomp}
\usepackage{xcolor}
\def\BibTeX{{\rm B\kern-.05em{\sc i\kern-.025em b}\kern-.08em
    T\kern-.1667em\lower.7ex\hbox{E}\kern-.125emX}}

\usepackage{bm,marvosym,mathtools,environ,array,amsthm,url}
\usepackage[ruled,vlined,linesnumbered,resetcount]{algorithm2e}

\graphicspath{{./figs/}}

\renewcommand{\v}[1]{\boldsymbol{\mathbf{#1}}} 
\newcommand{\opn}[1]{\operatorname{#1}} 
\newcommand{\set}[1]{\mathbb{#1}} 
\newcommand{\norm}[2]{\lVert #1 \rVert_{#2}} 
\newcommand{\tm}[0]{\times} 
\newcommand{\prt}[1]{\left(#1\right)} 
\newcommand{\sqb}[1]{\left[#1\right]} 

\newcommand{\inc}[1]{\in \set{C}^{#1}} 

\newtheorem{definition}{Definition}

\makeatletter
\newsavebox\myboxA
\newsavebox\myboxB
\newlength\mylenA
\newcommand*\xov[2][0.75]{%
    \sbox{\myboxA}{$\m@th#2$}%
    \setbox\myboxB\null
    \ht\myboxB=\ht\myboxA%
    \dp\myboxB=\dp\myboxA%
    \wd\myboxB=#1\wd\myboxA
    \sbox\myboxB{$\m@th\overline{\copy\myboxB}$}
    \setlength\mylenA{\the\wd\myboxA}
    \addtolength\mylenA{-\the\wd\myboxB}%
    \ifdim\wd\myboxB<\wd\myboxA%
       \rlap{\hskip 0.5\mylenA\usebox\myboxB}{\usebox\myboxA}%
    \else
        \hskip -0.5\mylenA\rlap{\usebox\myboxA}{\hskip 0.5\mylenA\usebox\myboxB}%
    \fi}
\makeatother

\NewEnviron{meq}{
    \begin{align}
        \begin{split}
            \BODY
        \end{split}
    \end{align}
    }

\begin{document}

\title{\LARGE Analysis of Partially-Calibrated Sparse Subarrays for Direction Finding with Extended Degrees of Freedom\\

\thanks{Funding: Pontifical Catholic University of Rio de Janeiro and CNPq (Brazil).}
}

\author{\IEEEauthorblockN{Wesley S. Leite}
\IEEEauthorblockA{\textit{Center for Telecommunications Research (CETUC)} \\
\textit{Pontifical Catholic University of Rio de Janeiro (PUC-Rio)}\\
Rio de Janeiro, Brazil \\
wleite@ieee.org}
\and
\IEEEauthorblockN{Rodrigo C. De Lamare}
\IEEEauthorblockA{\textit{Center for Telecommunications Research (CETUC)} \\
\textit{Pontifical Catholic University of Rio de Janeiro (PUC-Rio)}\\
Rio de Janeiro, RJ\\
delamare@puc-rio.br}

}

\maketitle

\begin{abstract}
This paper investigates the problem of direction-of-arrival (DOA) estimation using multiple partially-calibrated sparse subarrays. In particular, we present the Generalized Coarray Multiple Signal Classification (GCA-MUSIC) DOA estimation algorithm to scenarios with partially-calibrated sparse subarrays. The proposed GCA-MUSIC algorithm exploits the difference coarray for each subarray, followed by a specific pseudo-spectrum merging rule that is based on the intersection of the signal subspaces associated to each subarray. This rule assumes that there is no a priori knowledge about the cross-covariance between subarrays. In that way, only the second-order statistics of each subarray are used to estimate the directions with increased degrees of freedom, i.e., the estimation procedure preserves the coarray Multiple Signal Classification and sparse arrays properties to estimate more sources than the number of physical sensors in each subarray. Numerical simulations show that the proposed GCA-MUSIC has better performance than other similar strategies.\\
\end{abstract}


\section{Introduction} 
Recently, the field of direction-of-arrival (DOA) estimation with arrays of sensors have been largely investigated in applications involving sonar, radar and communications \cite{VanTrees2002,Zhang2013,jidf,Pal2010,sjidf}. In this sense, a great deal of research have been dealing with sparse sensor arrays techniques due to their remarkable performance improvements in beamforming applications \cite{sjidf,l1stap,lrcc}, as well as direction-of-arrival (DOA) estimation \cite{Chowdhury2021,Zachariah2017,Liu2016,Leite2021,Pal2010-1,Qiu2016}. In this context, the capability of sparse arrays to recover more sources than the number of physical sensors is a huge advantage of this kind of arrays over the traditional uniform linear array (ULA), even if they require some additional processing \cite{Chen2021,Liu2016,Liu2016-2,Leite2021,Pal2010-1,Pal2012,rdcoprime,misc22,emisc}. Extensions to subarrays applications have been developed, since they have the potential of reducing the communication overhead in the central processing unit and allow for parts of the array to be located in multiple platforms operating with asynchronous sampling schemes, i.e., partially-calibrated schemes. \cite{Gu2011,Elbir2019,See2004,Swindlehurst2001,Rieken2004,dfsubarrays}.

Among many signal processing strategies that have been developed to tackle the problem of estimating the directions-of-arrival (DOAs) with sparse arrays, one of the most used is the so-called Spatial Smoothing Multiple Signal Classification (SS-MUSIC) or coarray Multiple Signal Classification, in which spatial smoothing is used to build-up the rank of a matrix obtained from the difference coarray transformation \cite{Pal2010-1}. This technique has demonstrated to possess good resolution performance and exploitation of the array degrees of freedom. On the other hand, a subspace based approach designed to deal with partially-calibrated arrays generalized MUSIC to allow the DOA estimation in this context \cite{Rieken2004}. While the former was developed for coherent arrays, the latter significantly reduces the number of sources that can be estimated for a given amount of sensors.

In this work, we investigate the problem of DOA estimation using multiple partially-calibrated sparse subarrays. In particular, we present the Generalized Coarray MUSIC (GCA-MUSIC) DOA estimation algorithm that has all the good properties of SS-MUSIC and G-MUSIC. The proposed GCA-MUSIC algorithm consists of the exploitation of the difference coarray for each subarray, followed by a specific pseudo-spectrum merging rule that is based on the intersection of the signal subspaces associated to each subarray. This rule assumes that there is no a priori knowledge about the cross-covariance between subarrays. In that way, only the second-order statistics of each subarray are used to estimate the directions with increased degrees of freedom, i.e., the estimation procedure preserves the coarray MUSIC and sparse arrays properties to estimate more sources than the number of physical sensors in each subarray. Numerical simulations show that the proposed GCA-MUSIC algorithm has better performance than other similar strategies.

\emph{Paper structure}: In Section~\ref{sec:systemModel}, the system model and problem statement are presented. In Section~\ref{sec:propMethod}, the proposed GCA-MUSIC DOA estimation algorithm for sparse subarrays is detailed. In Section~\ref{sec:simulation}, the numerical results are used to demonstrate the performance of the proposed GCA-MUSIC algorithm, whereas Section~\ref{sec:conclusion} draws the conclusions.

\emph{Notation}: $\set{S}$, $a$, $\v{a}$ and $\v{A}$ indicate sets, scalars, column vectors, and matrices, respectively. $\opn{blkdiag}(\cdot)$ is the block diagonal matrix, whereas $\opn{colspan}(\v{A})$ represents the column space of $\v{A}$. 

\section{System Model and Problem Statement}\label{sec:systemModel}

The representation of the data acquisition model with partially calibrated subarrays for multiple snapshots is given by 
\begin{equation}\label{eq:mmv_model_dense}
    \v{x}^{(l)}(t) = e^{-j\phi_l}\v{A}_{\set{S}_l}(\v{\theta})\v{s}(t)+\v{n}_{\set{S}_l}(t)
\end{equation}
where $l=1,\ldots,L$, $t=1,\ldots,T$, $\phi_l$ is the $l$-th subarray phase shift, $\v{A}_{\set{S}_l}(\theta)\inc{N_l\tm D}$ is the $l$-th subarray manifold with the geometry defined by the set of integers $\set{S}_l$ (normalized positions in terms of $d$ - minimum intersensor spacing), the $l$-th subarray has $N_l$ sensors, and there are $D$ impinging sources with normalized directions given by $\v{\theta}\in [-1,1)^{D}$ (sine of DOAs - spatial frequency), $\v{s}\inc{D}$ is the source signal, $\v{X}^{(l)}\inc{N_l\tm T}$ is the $l$-th subarray received signal snapshots matrix, and $\v{n}_{\set{S}_l}(t)$ is the subarray noise snapshots matrix. The noise and the source signal are drawn from a zero-mean circularly complex multivariate Gaussian distribution. The noise is spatially white and the sources are assumed uncorrelated. Remark: as abuse of notation, we refer to the set $\set{S}$ defining the sensors locations as the array itself.

The set notation emphasizes the dependence of the equations on the sparse subarrays geometries denoted by $\set{S}_l$. The problem that we would like to solve is to find the normalized directions $\v{\theta}$, where both $\v{s}(t)$ and $\phi_l$ are unknowns (source signals and phase delay between subarrays). 

\section{Proposed GCA-MUSIC DOA Estimation Method}\label{sec:propMethod}

In this section, we present the GCA-MUSIC DOA estimation algorithm to partially calibrated array scenarios which extends the coarray MUSIC algorithm developed in \cite{Pal2010-1}. Unlike compressive sensing and sparsity-aware techniques \cite{Tirer2021,saalt,memd,dce,l1cgstap,jidf,jidfecho,sjidf,sbstap,damdc} that are very effective for scenarios with short data records but have a performance that might be far from the Cramer-Rao lower bounds (CRLBs), subspace techniques \cite{jio,mcg,wlmwf,wljio,jiostap,jiodf,locsme,Qin2015,okspme,lrcc,mskaesprit} can perform close to CRLBs when the statistics of the sensor data are accurately estimated. In this case, each of the subarrays has coherent sensors and possesses a sparse geometry. To this end, we recover the concepts and properties associated to Type-II Sparse Linear Arrays examined in \cite{Leite2022}. 

\subsection{Type-II Sparse Linear Arrays}

The so-called Type-II Sparse Linear Arrays corresponds to a union of subarrays, each of them with a predefined sparse linear geometry, denoted by $\set{S}_l$, $l=1,\ldots,L$ (L subarrays). Indeed, one of the key results presented in \cite{Leite2022} is that the number of degrees of freedom (DoF) of the entire array $\set{S}$ and the number of DoF of each of the subarrays $\set{S}_l$, denoted by $\text{sDoF}$, are related by $\text{DoF} \leq L(\text{sDoF}-1)+2(L-1)\mu+1$, for $1\leq \mu\leq\kappa$, where $\mu$ is the normalized distance between subarrays (in terms of the minimum intersensor distance $d$) and $\kappa$ is the subarrays individual aperture. For $\mu>\kappa$, $\text{DoF} = (2L-1)\text{sDoF}$. Those aspects will be exploited in the discussion that follows, which describes the proposed DoA estimator.

\subsection{Coarray MUSIC with Sparse Subarrays}

Since we are dealing with sparse subarrays, a natural starting point to develop an estimator would be to extend the coarray MUSIC (CA-MUSIC) algorithm \cite{Pal2010-1}, which we will call Generalized Coarray MUSIC (GCA-MUSIC). This estimation procedure has many advantages over other techniques presented in the literature: it is capable of exploiting half of the DoF of the difference sub-coarrays, presents super-resolution performance capabilities and has a reasonable trade-off in terms of computational burden. The proposed GCA-MUSIC aims to tackle the case of DoA estimation with partially calibrated sparse subarray geometries. To this end, we start by computing the second-order statistics associated to each of the subarrays, according to 
\begin{equation}\label{eq:cov_mat_sub}
    \v{R}_{\set{S}_l} = \v{A}_{\set{S}_l}(\v{\theta})\v{R}_{ss}\v{A}_{\set{S}_l}^H(\v{\theta})+\sigma_n^2 \v{I}\text{, } l=1,\ldots,L
\end{equation}
where $\v{R}_{\set{S}_l}$ is the received signal covariance matrix of the $l$-th subarray and $\v{R}_{ss}=\opn{diag}\prt{\sigma_1^2,\ldots,\sigma_D^2}$ is the uncorrelated sources covariance matrix. By vectorizing (\ref{eq:cov_mat_sub}), we arrive at 
\begin{equation}
    \v{z}_l = (\v{A}_{\set{S}_l}^{\ast}\circ\v{A}_{\set{S}_l})\v{p}+\sigma_n^2\bar{\v{i}}
\end{equation}
where $\circ$ denotes the Khatri-Rao product, $\bar{\v{i}}=\sqb{\v{e}_1^T,\ldots,\v{e}_N^T}^T$ is the vectorization of the identity matrix, and $\v{p}=\sqb{\sigma_1^2,\ldots,\sigma_D^2}$ contains the sources powers. Notice that to simplify the equations we adopted the same number of sensors (same aperture) for each of the subarrays (total of $NL$ sensors for the whole array). By removing the repeated rows in $\v{A}_{\set{S}_l}^{\ast}\circ\v{A}_{\set{S}_l}$ after their first occurrence (mirroring the operation in $\v{z}_l$ and $\bar{\v{i}}$) and sorting the virtual sensors (coarray) elements in ascending order, we have
\begin{equation}\label{eq:sca}
    \v{x}_{\set{D}_l}=\v{A}_{\set{D}_l}\v{p}+\sigma_n^2\v{i}
\end{equation}
where $\set{D}_l$ denotes the difference coarray set associated to the $l$-th subarray, $\v{x}_{\set{D}_l}\inc{|\set{D}_l|}$ is the $l$-th coarray received signal and $\v{i}\in\{0,1\}^{|\set{D}_l|}$ is an all-zero vector with the exception of a $1$ in its half position (element $(|\set{D}_l|+1)/2$).  

From this point, we introduce some terminology according to the following definitions:
\begin{definition}[Sparse Subarray (SpSub)]
    The sparse subarrays are defined as each partially calibrated part of the whole array. The sensors are coherent within each sparse subarray (sampling process is performed in a synchronized basis). They are denoted by $\set{S}_l$, with $l=1,\ldots,L$.
\end{definition}
\begin{definition}[Subcoarray (SCA)]
    A Subcoarray is defined as the Difference Coarray associated to each SpSub. They are denoted by $\set{D}_l$, with \emph{$l=1,\ldots,L$}.
\end{definition}
\begin{definition}[Spatially Smoothed Subcoarray (SS-SCA)]
    Spatially Smoothed Subcoarray (SS-SCA) is a SCA with reduced dimension dictated by the parameter choices of a spatial smoothing-like procedure. They are denoted by \emph{$\set{D}_l^i$}, with \emph{$l=1,\ldots,L$} and $i=1,\ldots, M$. 
\end{definition}

Notice that each SCA (associated to a specific SpSub), generated after the mathematical procedure described from (\ref{eq:cov_mat_sub}) to (\ref{eq:sca}), will have a total of $M$ SS-SCA. Then, we have a total of $M\cdot L$ SS-SCA for the whole array.   

We consider SpSub and respective SCA with central contiguous part (virtual ULA) large enough to allow a recovery of all of the sources DoA. To simplify the equations, we will assume that the SpSub has a filled SCA (no holes in virtual domain), i.e., the second-order statistics associated to each SpSub contains all the correlation lags from 0 up to $\kappa = (|\set{D}_l |-1)/2$ (the aperture of each SpSub is equal between all the subarrays, because we are considering Type-II Arrays with the same number of physical sensors within each SpSub).

By resorting to the rank properties, it is clear that the outer product $\v{x}_{\set{D}_l}\v{x}_{\set{D}_l}^H$ is rank deficient. Then, we build up this rank using $M=\kappa+1=(\text{sDoF}+1)/2$ SS-SCA (forward spatial smoothing), for each SCA/SpSub, according to
\begin{equation}\label{eq:ss_eq}
    \v{R}_{\set{D}_l}^{\text{SS}} =\frac{1}{M}\sum_{i=1}^{M}\v{x}_{\set{D}_l^i}\prt{\v{x}_{\set{D}_l^i}}^H
\end{equation}
where $\v{x}_{\set{D}_l^i}\inc{M}$ is the $i$-th overlapping SS-SCA of the $l$-th SCA, starting ($i=1$) from the maximum value of the contiguous part of the SCA, and $\text{sDoF}=|\set{D}_l|$ is the number of degrees of freedom for each subarray. We remark that although many spatial smoothing techniques can be used in (\ref{eq:ss_eq}) \cite{80975}, depending on the amount of computational resources available in the DSP, we keep the standard SS as presented in \cite{Pal2010-1} because it resulted in a good estimation performance in our numerical results. Additionally, it can be demonstrated that $\eqref{eq:ss_eq}$ has a signal and noise subspace that allows us to obtain the sources DoA by using MUSIC. Then, each $\v{R}_{\set{D}_l}^{\text{SS}}$, originated from each of the partially calibrated subarrays provides rough estimates of the sources DoA.

The second problem we deal with is how to combine the processing such that we can profit from the estimates of each SpSub in an integrated fashion. This spectrum combination is key to increasing the estimation performance, as will be demonstrated further.

To perform the signal decomposition, we adopt a similar strategy as described in \cite{Rieken2004}. The signal and noise subspace of $\v{R}_{\set{D}_l}^{\text{SS}}$ can be obtained from the following EVD
\begin{equation}
    \v{R}_{\set{D}_l}^{\text{SS}} = 
    \begin{bmatrix}
        \v{U}_{l}& \v{V}_{l}
    \end{bmatrix}
    \opn{diag}\prt{\beta_{1}^l,\ldots,\beta_{N}^l}
    \begin{bmatrix}
        \v{U}_{l}^H\\ \v{V}_{l}^H
    \end{bmatrix}
\end{equation}
that has the same eigenvectors (signal and null-space) as those associated to the first array manifold of the spatial smoothing procedure, denoted by $\v{A}_{\set{D}_l^1}$ (the virtual array manifold corresponding to the last $M$ rows of $\v{A}_{\set{D}_l}$). 

Particularly in this case, the subarrays are not coherent and so the statistics are divided and must be processed separately at some degree. The coarray received signal can be written in terms of the SCA received signal, after the dimensionality reduction imposed by the spatial smoothing technique. Mathematically, following the strategy described in \cite{Rieken2004}, based on the method of projection onto convex sets, we can write
\begin{equation}
    \Tilde{\v{U}}_l = \opn{blkdiag}\prt{\v{I}_{N(l-1)},\v{U}_l,\v{I}_{N(L-l)}}
\end{equation}
where $\opn{colspan}(\Tilde{\v{U}}_l)$ corresponds to the signal subspace associated to the $l$-th subarray. Since this matrix is orthonormal, then its projection matrix $\v{P}_{\Tilde{\v{U}}_l}=\Tilde{\v{U}}_l\Tilde{\v{U}}_l^H$ serves as a proxy to the so-called synthetic signal subspace, that is an intersection of $\opn{colspan}(\Tilde{\v{U}}_l)$ for all $L$. By iterating with the method of projection onto convex sets, we can demonstrate that the projection onto the intersection of the subspaces generated by $\opn{colspan}(\Tilde{\v{U}}_l)$ is given by
\begin{equation}\label{eq:signal_cal_subs}
    \v{P} = \opn{blkdiag}\prt{\v{U}_1\v{U}_1^H,\ldots,\v{U}_L\v{U}_L^H}
\end{equation}
which implies that the signal subspace of all the partially calibrated subarrays is a subset of 
\begin{equation}
    \opn{colspan}(\opn{blkdiag}(\v{U}_1,\ldots,\v{U}_L))=\opn{colspan}(\v{U})
\end{equation}
Thus, the kernel associated to (\ref{eq:signal_cal_subs}) is then given by $\mathcal{N}(\v{U}^H) = \v{I}-\v{P}$, that can be used to find the DoAs using the steering vector of the whole array.

Lastly, we would like to point out that GCA-MUSIC is capable of identifying more sources than sensors even for the partially calibrated array scenario. In what follows, we perform a comparison between our approach and the one developed in \cite{Rieken2004}, which is termed here G-MUSIC. Indeed, assuming each subarray has $N$ physical sensors (total of $LN$ sensors), G-MUSIC can identify only up to $N-1$ sources, since it computes the pseudo-spectrum for each of the subarrays using conventional MUSIC in standard domain. 

On the other hand, GCA-MUSIC can identify up to $(\text{sDoF}-1)/2$ sources. To write this quantity as a function of the number of physical sensors, we must define the geometry we are dealing with. For example, for sparse subarrays following a Two-Level Nested Array geometry, GCA-MUSIC can identify up to $N^2/2+N/2-1$ sources, which is the same number of sources that coarray MUSIC can identify with a coherent array of $N$ sensors with this geometry. Then, GCA-MUSIC takes advantage of the correlation lags to estimate much more sources than G-MUSIC, with the advantage of maintaining the super-resolution properties associated to eigenspace-based DOA estimation algorithms.

The overall complexity of coarray MUSIC is $\mathcal{O}(TM+M^3)$ \cite{Liu2015}, that is dominated by the eigen-decomposition of $\v{R}_{\set{D}}^{\text{SS}}$. Following that, our GCA-MUSIC approach has an overall complexity of $\mathcal{O}(TL^2M+LM^3)$, where $M$ is the number of SS-SCA for each SpSub and $T$ is the total number of snapshots for each SpSub.

\section{Simulation}\label{sec:simulation}

In this section, we evaluate the GCA-MUSIC performance capabilities through numerical experiments. The GCA-MUSIC algorithm performance is compared to the Generalized MUSIC and a modified SS-MUSIC algorithm which combines the pseudo-spectra by simple averaging them  \cite{Pal2010-1,Rieken2004}. In order to do that, two scenarios are used: the first one with less sources than sensors and an alternative scenario with more sources than sensors, that is one of the key advantages of using sparse subarrays with Khatri-Rao product-based processing. To this end, we employ Uniform Linear Arrays (ULA), Two-level Nested Arrays (NAQ2), Minimum Redundancy Arrays (MRA) and Second-Order Super Nested Arrays (SNAQ2) \cite{VanTrees2002,Pal2010-1,Moffet1968,Liu2016}. 

The simulation scenarios adopt $L=3$ sparse subarrays with $N=7$ sensors each,  $d/\lambda=1/2$, and $\mu=1$. The performance curves are drawn by assessing the Root Mean Square Error (RMSE) \cite{VanTrees2002} $\textrm{RMSE}=\sqrt{\frac{1}{DR}\sum_{i=1}^{R}\norm{\v{\theta}-\hat{\v{\theta}}_i}{2}^2}$.
We also add that we use $R=1000$ Monte Carlo runs to have well behaved curves and the phase shifts $\v{\phi}_l$ are drawn from $U(0,2\pi)$ for each subarray and run. The amount of data for the SNR curves is $T=100$ snapshots.

\subsection{Arrays and subarrays characterization}
\begin{figure}[htbp]
\centering
\includegraphics[width=.47\textwidth]{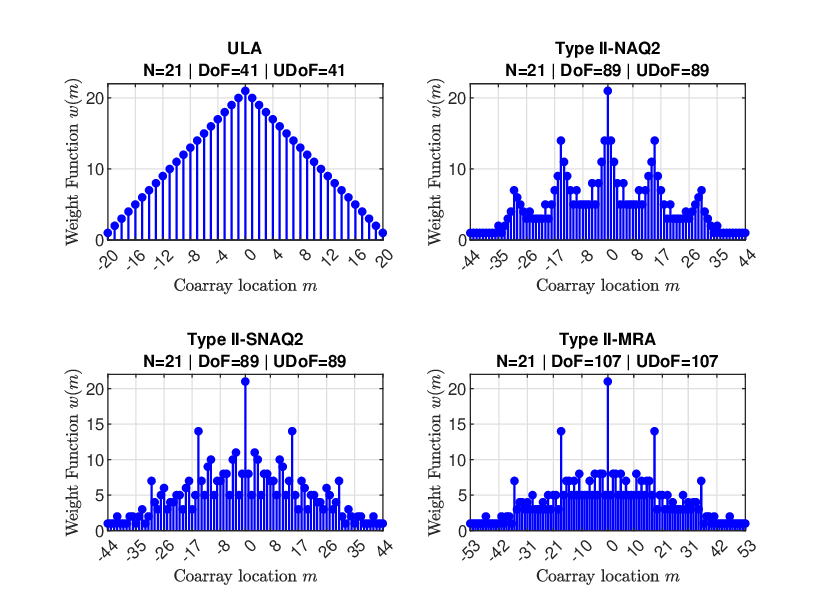}
\caption{Weight functions for Type-II arrays with $L=3$ subarrays and $N=7$ sensors each.}
\label{fig:0}
\end{figure}

\begin{figure}[htbp]
\centering
\includegraphics[width=.47\textwidth]{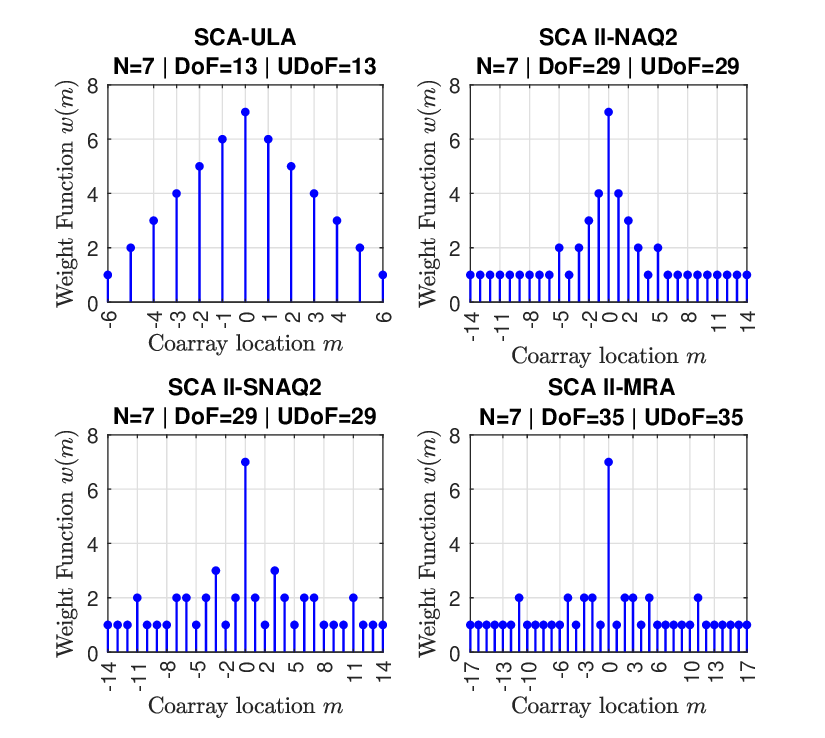}
\caption{Weight functions for each SpSub and the associated SCA with $N=7$ sensors.}
\label{fig:1}
\end{figure}

Fig.~\ref{fig:0} shows a comparison between the degrees of freedom for the different geometries employed in this section. Notice that the Type II-MRA geometry has a filled coarray with $\text{DoF}=107$, followed by NAQ2 and SNAQ2 with $\text{DoF}=89$, and the ULA with $\text{DoF}=41$. The structures of the SCA associated to each of the geometries are illustrated in Fig.~\ref{fig:1}, which shows the number of DoF for the subarray of each geometry. As it was expected, MRA SpSub possesses more DoF, followed by NAQ2, SNAQ2 and ULA. Notice that we could calculate the number of DoF of the whole array by means of the upper bound $\text{DoF}=L(\text{sDoF}-1)+2(L-1)\mu+1$, provided in \cite{Leite2022}, which becomes an equality because the SCAs have no holes. For example, for SNAQ2, $\text{DoF}=3(29-1)+2(3-1)\cdot 1+1=89$.  

\subsection{GCA-MUSIC performance comparison}

In this section, we compare the performance of GCA-MUSIC, G-MUSIC, and AVCA-MUSIC, that consists of a version of coarray MUSIC where we average the pseudo-spectrum obtained with $\v{R}_{\set{D}_l}^{\text{SS}}$ for each SCA. 

\begin{figure}[htbp]
\centering
\includegraphics[width=.47\textwidth]{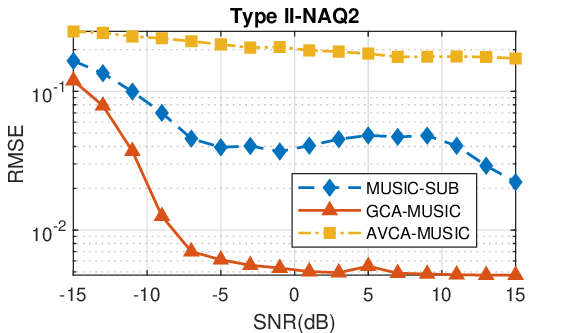}
\caption{RMSE performance curves against SNR for Type-II NAQ2 array. $\text{T}=100$ snapshots. $D=6$ sources located at $\v{\theta}=[-0.7,-0.5,-0.3,0.3,0.5,0.7]$.}
\label{fig:2}
\end{figure}

Fig.~\ref{fig:2} shows the RMSE against the SNR with $D=6$ sources. Clearly, GCA-MUSIC presents a much better performance in comparison to the other algorithms. This performance gain becomes more prominent as we increase the SNR. We also note that in contrast to the results in \cite{Rieken2004}, where the averaging of the individual spectrum results in almost the same performance in source localization, GCA-MUSIC relies heavily on the intersection of the subspaces to increase the estimation accuracy and the intuitive spectrum averaging has a poor performance. 

\begin{figure}[htbp]
\centering
\includegraphics[width=.47\textwidth]{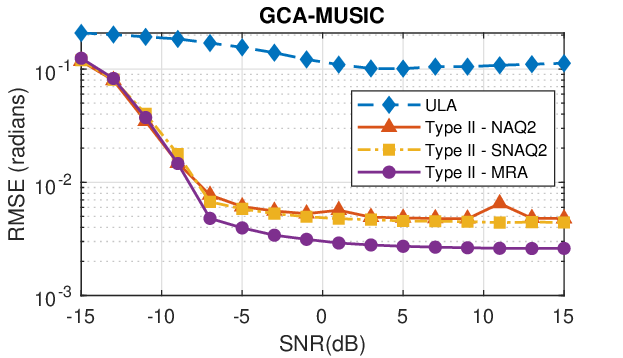}
\caption{RMSE performance curves against SNR for GCA-MUSIC with four geometries: ULA, NAQ2, SNAQ2, and MRA. $\text{T}=100$ snapshots. $D=6$ sources located at $\v{\theta}=[-0.7,-0.5,-0.3,0.3,0.5,0.7]$.}
\label{fig:3}
\end{figure}

Fig.~\ref{fig:3} shows a comparison among the different arrays against the SNR for GCA-MUSIC and $D=6$ sources. It is clear that this algorithm performs better with MRA, followed by SNAQ2 and NAQ2. The ULA geometry presents the worst performance, which is justified by its smaller aperture and smaller number of DoF for each SCA.  

\begin{figure}[htbp]
\centering
\includegraphics[width=.47\textwidth]{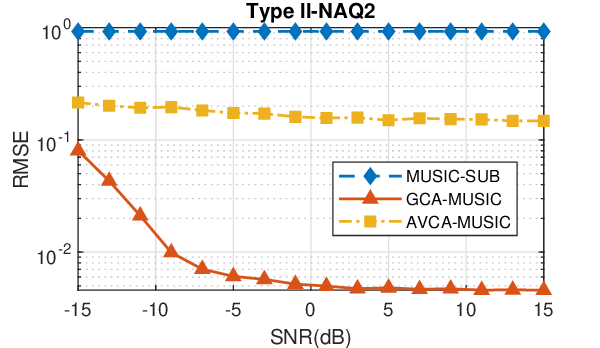}
\caption{RMSE performance curves against SNR for Type-II NAQ2 array. $\text{T}=100$ snapshots. $D=6$ sources located at $\v{\theta}=[-0.7,-0.5,-0.3,0,0.3,0.5,0.7]$.}
\label{fig:4}
\end{figure}

Fig.~\ref{fig:4} shows the RMSE against the SNR, but this time in a scenario with more sources that the number of individual sensors in each subarray ($D=7$ sources and $L=3$ subarrays with $N=7$ sensors each). As expected, the estimation performance for G-MUSIC degrades largely, as it is capable of estimating only up to $N-1=6$ sources. This justifies the superiority of the proposed GCA-MUSIC algorithm for partially calibrated sparse arrays. 

\section{Conclusion}\label{sec:conclusion}
In this paper, we have investigated partially calibrated sparse arrays and their use to perform DOA estimation. We have devised a new algorithm called GCA-MUSIC that extends the coarray MUSIC algorithm to partially-calibrated sparse arrays. The  GCA-MUSIC algorithm is capable of estimating more sources than sensors and can estimate the DoAs with very good performance in comparison to its counterparts when using this kind of noncoherent processing. 

\bibliographystyle{IEEEtran}
\bibliography{mybib}

\end{document}